\documentclass{article}

\usepackage{arxiv}

\usepackage[utf8]{inputenc} 
\usepackage[T1]{fontenc}    
\usepackage{hyperref}       
\usepackage{url}            
\usepackage{booktabs}       
\usepackage{amsfonts}       
\usepackage{nicefrac}       
\usepackage{microtype}      
\usepackage{lipsum}		
\usepackage{graphicx}
\usepackage{natbib}
\usepackage{doi}
\usepackage{tabularx}
\usepackage{graphicx}
\usepackage{caption}
\usepackage{subcaption}
\usepackage{tabularx, multirow, booktabs}

\usepackage{times}
\usepackage{latexsym}

\usepackage[T1]{fontenc}

\usepackage[utf8]{inputenc}

\usepackage{microtype}
\usepackage{inconsolata}
\usepackage{graphicx}
\usepackage{dblfloatfix}
\usepackage{siunitx} 
\usepackage{booktabs} 
\usepackage{tabularx} 
\usepackage{booktabs}  
\usepackage{multirow} 
\usepackage{siunitx}

\title{Efficiently Leveraging Linguistic Priors for Scene Text Spotting}

\date{}                     

\author{
  Nguyen Nguyen \\
  University of Rochester\\
  Rochester, NY, USA \\
  \texttt{nguyen.nguyen@rochester.edu} \\
  \And
  Yapeng Tian \\
  University of Texas at Dallas\\
  Richardson, TX, USA \\
  \texttt{yapeng.tian@utdallas.edu} \\
  \And
  Chenliang Xu \\
  University of Rochester\\
  Rochester, NY, USA \\
  \texttt{chenliang.xu@rochester.edu} \\
}

\renewcommand{\shorttitle}{Efficiently Leveraging Linguistic Priors for Scene Text Spotting}

\hypersetup{
pdftitle={OSCaR: Object State Captioning and State Change Representation},
pdfsubject={q-bio.NC, q-bio.QM},
pdfauthor={Nguyen Nguyen, Jing Bi, Ali Vosoughi, Yapeng Tian, Pooyan Fazli, Chenliang Xu},
pdfkeywords={First keyword, Second keyword, More},
}

\begin{document}
\maketitle

\begin{figure}[h]
  \centering
  \includegraphics[width=0.6\textwidth]{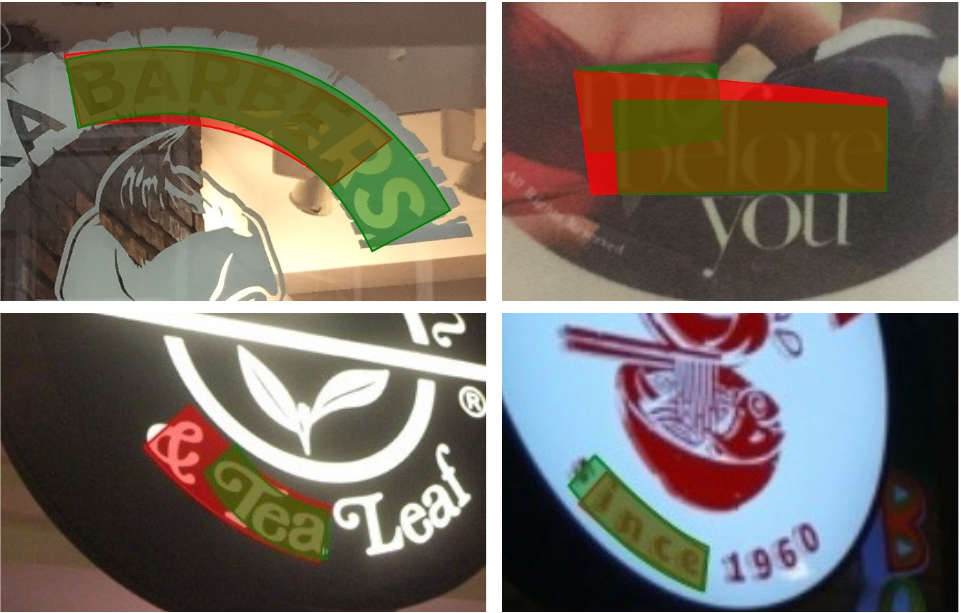}
  \caption{Comparison of detection results \textbf{with} ({\color{green} green shaded}) and \textbf{without} ({\color{red} red shaded}) language knowledge prior guidance. Language prior is not only helpful for text recognition but also for text detection.}
  \label{fig:example}

\end{figure}

\begin{abstract}
Incorporating linguistic knowledge can improve scene text recognition, but it is questionable whether the same holds for scene text spotting, which typically involves text detection and recognition. This paper proposes a method that leverages linguistic knowledge from a large text corpus to replace the traditional one-hot encoding used in auto-regressive scene text spotting and recognition models. This allows the model to capture the relationship between characters in the same word. Additionally, we introduce a technique to generate text distributions that align well with scene text datasets, removing the need for in-domain fine-tuning. As a result, the newly created text distributions are more informative than pure one-hot encoding, leading to improved spotting and recognition performance. Our method is simple and efficient, and it can easily be integrated into existing auto-regressive-based approaches. Experimental results show that our method not only improves recognition accuracy but also enables more accurate localization of words. It significantly improves both state-of-the-art scene text spotting and recognition pipelines, achieving state-of-the-art results on several benchmarks.

\end{abstract}

\footnote{This work was supported in part by NSF under Grant 1909912, NIH under Grant R01EY034562, DARPA under Contract HR00112220003, and the Center of Excellence in Data Science, an Empire State Development-designated Center of Excellence. The content of the information does not necessarily reflect the position of the Government, and no official endorsement should be inferred.}

\section{Introduction}
Text detection and recognition in natural scenes is a research area of considerable importance, with diverse practical applications ranging from aiding the visually impaired \cite{Jabnoun2017ANM} to facilitating robot navigation \cite{Liu2021ARO} and enabling mapping, and localization \cite{Greenhalgh2015RecognizingTT}. However, many text instances in natural settings exhibit inherent ambiguity stemming from aesthetic variations, environmental deterioration, or poor illumination conditions. This ambiguity can often be partly reduced by considering a list of lexicons or a dictionary \cite{Nguyen2021DictionaryguidedST}. However, this usually increases the complexity of models.

An end-to-end text spotting system typically involves two main steps: text detection and text recognition. Over the years, researchers have devoted considerable effort to advancing the state of the arts in both text detection~\cite{Baek2019CharacterRA, tang2022optimal, long2018textsnake, zhang2020deep, zhu2021fourier, liao2020real, liu2019pyramid, wang2019panet} and text recognition~\cite{Xie2022TowardUW, Bautista2022SceneTR, Wang2022MultiGranularityPF, Zhao2022BackgroundInsensitiveST, Liu2022OpenSetTR, Bhunia2021TowardsTU}. Despite advancements, current scene text spotting models do not match human reading capabilities, especially in recognizing distorted or blurry characters. This remarkable feat is due to the fact that linguistic knowledge provides a powerful prior that can help disambiguate text. Existing autoregressive text recognition models \cite{shi2016robust, wang2017gated, cheng2018aon, chng2019icdar2019} can leverage some of the linguistic structure from the training data in the decoder component, where the model predicts the next character by using previous character predictions as inputs. However, these approaches have yet to fully use linguistic knowledge for the following reasons. Firstly, existing scene text datasets are limited, making effective linguistic knowledge capture challenging. Secondly, current models employ basic one-hot vector encoding, ignoring the uncertainty or relationship between characters. Furthermore, autoregressive models predict character distribution based on previous characters, contrasting traditional methods that align this with one-hot, which has no connections to other characters. Thus, better modeling of linguistic knowledge is needed in pipelines that rely solely on autoregressive models.

During inference, text detection models often fail if the distance between two characters of the same word is too far (detecting them as two different words), or if the distance between two different words is too close (detecting them as one word), or if the word length is too long (missing some characters).
Traditional methods use a one-hot label, which can encourage better learning in detection when considering recognition as a constraint for detection models. However, this method only looks at each character independently, without considering the context of the entire word. This limits the ability of the detection model to look at the entire word in a comprehensive way.
Linguistic priors allow detection models to look at the entire word, which helps to improve detection performance. 
For example, we know that words are typically made up of multiple characters, and that the characters in a word are typically spaced close together. By taking linguistic priors into account, detection models can better understand the context of the entire word and make more accurate predictions.

In this paper, we propose a novel end-to-end scene text spotting method that efficiently incorporates language priors from trained language models. Our approach can also be viewed as a multimodal knowledge distillation technique, which has yet to be fully utilized for scene text spotting. By directly leveraging language model output as label embeddings to guide text spotting learning, our approach provides more detailed guidance than traditional one-hot encoding.
We empirically tested our approach on multiple benchmark datasets, including Total-Text \cite{Chng2017TotalTextAC}, SCUT-CTW1500 \cite{Liu2019CurvedST}, and ICDAR2015 \cite{Karatzas2015ICDAR2C}, demonstrating its remarkable effectiveness in transferring knowledge from language models to scene text spotting models. Besides, to demonstrate the generalizability of our proposal, we also implemented our method on a state-of-the-art scene text recognition pipeline. Notably, our method allows for direct utilization of language model knowledge by scene text models without requiring any fine-tuning in the specific domain of scene text data. 

Our contributions are threefold:
 \begin{itemize}
\item We demonstrate the efficacy of leveraging language knowledge derived from a large language model to enhance the performance of both scene text spotting and recognition models. Specifically, we illustrate how the incorporation of language knowledge through the technique of knowledge distillation can benefit both text detection and text recognition tasks.
\item We present a novel and intuitive approach for integrating language models into the scene text spotting and recognition learning process without increasing model complexity in training and testing. Our proposed approach has been shown to be highly effective and is supported by empirical validation.
\item We introduce an innovative method that enables the utilization of pre-trained language model representations without the need for domain-specific fine-tuning on the scene text data, thus increasing the applicability and efficiency of the scene text spotting and recognition model.
 \end{itemize}

\begin{figure*}
    \centering
    \includegraphics[width=1.0\textwidth]{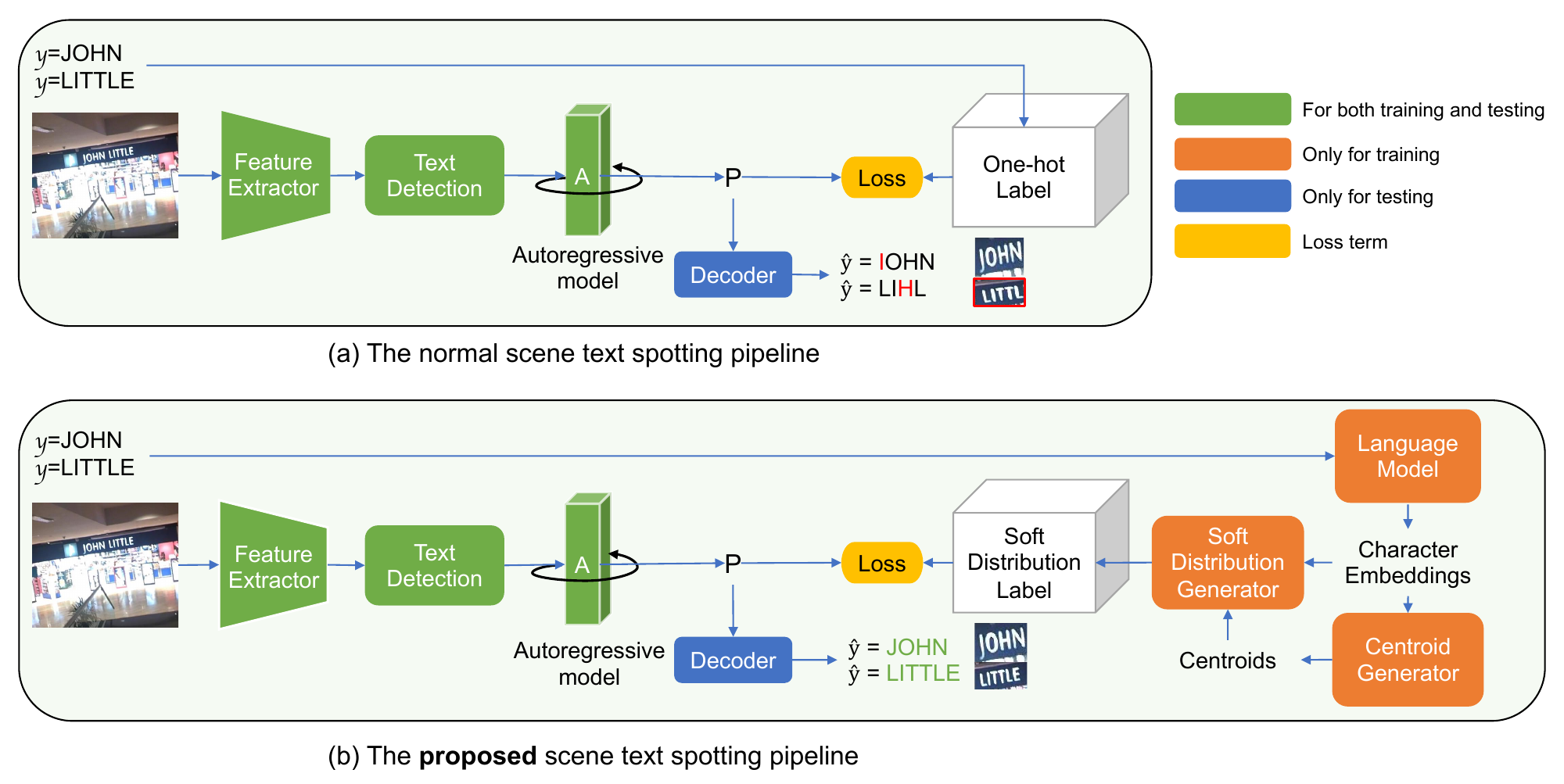}
    \vspace{-5mm}
    \caption{{\bf Traditional spotting pipeline (a) and proposed pipeline (b) on training.} In the traditional pipeline, models use the one-hot label directly to guide the training for the scene text system. Our proposal replaces the one-hot encoding by using soft distributions for every label character and improving detection and recognition results. Besides, we proposed a method to leverage knowledge from pretrained language models and construct the soft distribution well-adapted to the scene text domain without finetuning language models. \label{fig:recognition_pipeline}}
     \vspace{-5mm}
\end{figure*}

\section{Related Works}

\noindent \textbf{Scene Text Auto-regression Model.}
Auto-regressive models are widely used to tackle scene text recognition \cite{su2014accurate, shi2016robust, shi2016end, liu2016star, wang2017gated, cheng2018aon, chng2019icdar2019}. For example, CRNN \cite{shi2016end}   applied a deep bidirectional LSTM to infer the output probability distribution of the text. A CTC loss function removes duplicate characters in the output text series. In addition to CRNN, SCATTER \cite{litman2020scatter} sets up an attention layer to capture the intra-sequence relationships. Charnet \cite{liu2018char} fuses character-level and word-level encodings over the input to make the output more robust.
Taking good practice in past research, we adopt an auto-regressive method that decodes output sequences at the character level. A canonical RNN can execute the auto-regressive process with a unique direction from left to right.

\vspace{1mm}
\noindent \textbf{Language-driven for Scene Text Recognition.}
The integration of language knowledge into scene text recognition models has emerged as a popular research direction~\cite{Bautista2022SceneTR, Fang2021ReadLH, Da2022LevenshteinO, Wang2021FromTT}. For example, VisionLAN\cite{Wang2021FromTT} proposed a visual reasoning module that simultaneously captures visual and linguistic information by masking the input image at the feature level. Similarly, ABINet~\cite{Fang2021ReadLH} utilized a language model for iterative processing to enhance recognition accuracy with each iteration. Additionally, LevOCR~\cite{Da2022LevenshteinO} learned a refinement policy with insertion and deletion actions to iteratively enhance recognition accuracy. 
However, these approaches face two common challenges. Firstly, they increase the number of parameters in the models, which can lead to computational inefficiencies or slower model performance when iterative running is required. Secondly, these methods often rely on limited exploitation of language knowledge from small-scale scene text data without fully leveraging the vast knowledge from large-scale language models. ABINet++, an extension of ABINet, incorporates linguistic knowledge into scene text spotting models. However, like ABINet, this approach adds complexity through iterative execution, thereby elevating the model's complexity.
The utilization of language knowledge for scene text spotting remains relatively limited. To the best of our knowledge, our work represents the first attempt to leverage language knowledge from large language models without increasing model complexity or additional computational requirements during training and inference. 
This efficient approach can improve both text spotting and detection.

\noindent \textbf{Knowledge Distillation.}
Using knowledge transfer from a teacher to a student model can enhance performance at low cost \cite{hinton2015distilling}. Typically, the teacher's output probabilities guide the student model through a loss function optimization. DeiT \cite{Touvron2021TrainingDI} distinguishes between soft and hard-label distillation. Soft distillation directly uses the teacher's predictions, while hard-label distillation converts these to one-hot encoding. In this work, we propose a cross-modal knowledge distillation to transfer knowledge from text to image for scene text spotting.

\section{Language-guided Scene Text Spotting}
\label{sec:method}

We introduce a novel approach to enhance in-the-wild scene text spotting by integrating prior language knowledge. Scene text spotting locates text instances and computes their feature maps from an image. Using an auto-regressive model, we produce a probability map for each text instance, representing character distributions. We then compute embeddings for text labels and establish character prototypes. Using these embeddings and character prototypes, we create a soft distribution for each character. Ultimately, we obtain a distribution matrix for each word label, where each column represents a soft distribution corresponding to one character in the word. Subsequently, we employ these distributions as soft label distributions for the training process.

In this section, we describe the pipeline for scene text spotting, highlighting how the soft label distribution is created. Additionally, we discuss our loss function. Section Autoregressive-based Scene Text Recognition presents an in-depth description of the auto-regressive model for scene text recognition. Furthermore, in section Character Embedding, we delve into the topic of character embedding. Our approach for centroids estimation and soft distribution generation will be introduced below.

\subsection{Autoregressive-based Scene Text Recognition}
\label{sec:auto-based}

Auto-regressive models generate outputs over time, using past predictions as current inputs, described as $P(x_t|x_{t-1},...,x_{1})$. Traditional training uses one-hot labels to represent characters, where a character $c$ might be shown as $[0,0,1,...,0]$. However, this method fails to capture relationships between characters in natural language contexts. For example, it does not represent the likelihood of character $x$ following $b$ being less than $o$ after $b$. Although widespread in natural language, this knowledge cannot be effectively represented using one-hot encoding. In other words, during training, the model approximates a conditional distribution with a probability distribution to represent the character, irrespective of the context of the characters surrounding it.

The auto-regressive model used in training captures limited contextual knowledge due to scarce scene text datasets. Despite pre-training with synthetic datasets, the information learned is constrained by the dictionary used for scene text image generation. Moreover, during the fine-tuning process, the model tends to forget the limited knowledge captured from the pre-training data. Additionally, the recognition step of the scene text spotting pipeline is often implemented with lightweight models, restricting their ability to generalize with language. To address these challenges, we propose augmenting the scene text spotting model with a language model, which can extract knowledge from vast amounts of text data and enhance the model's ability to learn image patterns and language in a more comprehensive way.

\vspace{-1mm}
\subsection{Character Embedding}
\label{sec:char-emb}

Language models require converting text sequences into numeric embeddings through tokenization, either at the word or sub-word level. For tasks like scene text spotting, which prioritize learning character distributions, a character-level pre-trained model is more effective. CANINE~\cite{Clark2022CaninePA}, a recent model, differs from traditional tokenizers such as Byte Pair Encoding, WordPiece, and SentencePiece, as it employs a Transformer encoder trained directly on Unicode characters. Although this increases sequence length, CANINE addresses this issue with an efficient downsampling technique before applying the deep Transformer encoder. The CANINE model consists of a downsampling function, ENCODE, and an upsampling function, which operate on an input sequence of character embeddings e $\in$ $R^{n \times d}$  with length $n$ and dimension $d$, resulting in the output sequence embedding $\mathbf{Y}_{seq}$:
\begin{equation}
    \mathbf{Y_{seq}} \leftarrow \text{Up(ENCODE(DOWN(e)))}.
\end{equation}

\noindent To ensure that different characters do not have the same representation, the model uses a generalized hashing approach that concatenates the representations associated with various hash values. In our work, we utilize the pre-trained embedding $\mathbf{Y}_{seq}$ to estimate the centroid for each character, which is then considered the prototype for the embedding clusters.

\subsection{Centroid Generation}
\label{sec:centroid}

To generate a probability distribution from a given representation, a matrix multiplication operation is performed between the representation vector, denoted by $\textbf{x}$, and a weight matrix, denoted by $\textbf{W}$, which produces a vector of logit values. The logit vector is then fed through a softmax function, resulting in a probability distribution vector, denoted by $\textbf{D}$, which represents the probability of the representation belonging to each character in the vocabulary. The weight matrix $\textbf{W}$ is a prototype matrix, where each column corresponds to a prototype of a character in the vocabulary. The similarity between a representation and each character prototype in the latent space needs to be calculated in order to determine the most appropriate label for the representation. This similarity measurement allows for the selection of the label that is most compatible with the embedding.

\begin{equation}
     D_i = \frac{\exp(W_i^{T}x_j)}{\sum_{i=1}^{k} \exp(W_i^{T}x_j)},
     \label{eq:equ1}
\end{equation}
where $W_i$ is $i^{th}$ column of prototype matrix, $x_j$ is the representation of the character $j^{th}$ in the word, and $k$ is the size of character vocabulary.

In our approach, selecting an optimal \textbf{W} in Equation~(\ref{eq:equ1}) is crucial, where columns serve as character prototypes. Using the W-matrix directly from the language model might yield a too general probability distribution, causing disparity between scene text data and text data utilized to train the language model. Additionally, fine-tuning the language model with scene text data can introduce biases, compromising the model's generality. Furthermore, the process of fine-tuning the language model is time-consuming. In this paper, we propose using each prototype as the centroid of its character representation cluster by averaging all character representations of the same cluster:

\begin{equation}
    W_i = \frac{1}{n}\sum_{j=1}^{n} x_{ij},
    \label{eq:equ2}
\end{equation}
where $x_{ij}$ is the $j^{th}$ representation of character $i^{th}$, which is equivalent to $W_i$ in prototype matrix.

\begin{figure}
    \centering
    \includegraphics[width=0.6\textwidth]{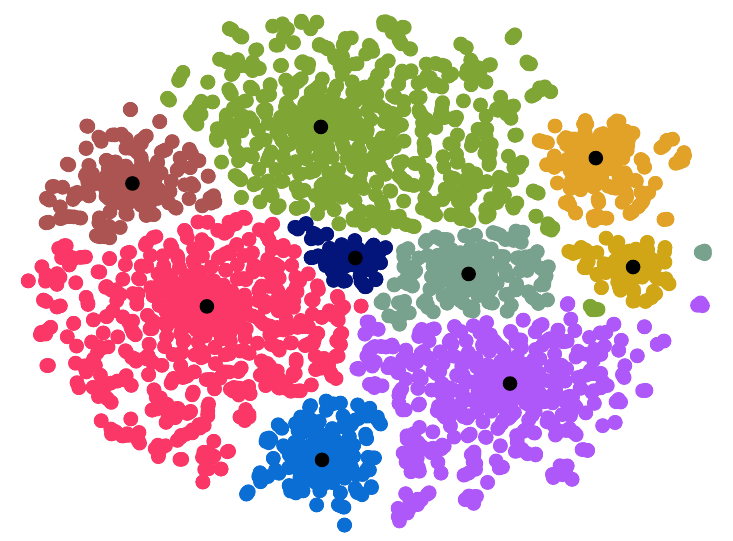}
    \caption{\textbf{Centroid Estimation.} Visualization of character embedding for 9 characters $a,b,c,d,e,f,g,h,i$. Each cluster equivalent with a character, and black points in the center are the centroids generated by ~(\ref{eq:equ2}).}
    \label{fig:centroid}
\end{figure}

\subsection{Soft Distribution Generation}
\label{sec:softdist}

Using the matrix \textbf{W} from equation \ref{eq:equ2}, we derive the soft label distribution for training from equation \ref{eq:equ1}. Our method bridges the gap between the pretrained language model and the scene text dataset without the need for fine-tuning. The texts we use to generate the $W$ matrix have the trade-off between scene text dataset and text from outside dataset. Basically, this is a sampling process to approximate the real centroids of language model. Using only the scene text dataset risks overfitting and losing general language knowledge. Conversely, using many external words keeps the centroids close to the original language model weights. In this work, we generate the centroids by a subset of words from a generic dictionary~\cite{Gupta2016SyntheticDF}, and the whole word appeared in the scene text training dataset. We will discuss more about our settings in the experiments part. The predicted distribution of the model will be denoted by distribution \textbf{P}. We use KL-Divergence as the loss function to measure the difference between the predicted distribution and our soft label distribution:

\begin{equation}
    KL(D||P) \propto - \sum_{i=1}^{k} D_{i}\log(P_{i}).
\end{equation}

Finally, we sum the loss of all columns together as our final loss function:
\begin{equation}
    \mathcal{L}(x,y) = \sum_{i=1}^{l} KL(D^{(i)}||P^{(i)}).
\end{equation}
where $D^{(i)}$ and $P^{(i)}$ are $i^{th}$ predicted distribution and soft label distribution, respectively. $l$ is the longest length over all words.

During the creation of the soft distribution label, we perform post-processing to remove noise. First, we check whether the distributions can represent mislabels by checking which character the highest probability belongs to. Next, distributions are filtered using a threshold $T$; those with a label probability greater than or equal to $T$ are retained. For those that don't meet the threshold, a predefined distribution is used. In positions with a one-hot representation of 1, we set the value to $T$, and all other positions are set to $P_i$:

\begin{equation}
    P_i = \frac{1-T}{k-1},
    \label{eq:thresh}
\end{equation}
where $T$ is the probability threshold we set for the label, $P_i$ is the probability value of the characters that are not label, and $k$ is the size of the character vocabulary.

The parameter $T$ greatly influences generated distributions. A low $T$ can yield overly flat distributions with high entropy, while a high $T$ may force shifts to non-linguistic-based ones. We determine $T$ using text labels. A 0.85 threshold effectively curbs excessive entropy while retaining language uncertainty. With this threshold, the error rate across datasets was satisfactory ($\leq$0.8\%), and any errors were post-processed. This post-processing is done once before training, removing the need for recalculations later. Our approach is straightforward, with the potential for future improvements in threshold selection.

\subsection{Implementation Details}

In our scene text spotting approach, we used the CANINE~\cite{Clark2022CaninePA} model to extract character representations from the dictionary. We then averaged embeddings of the same class from a vocabulary combining the dataset and 30k random words from a general dictionary~\cite{Gupta2016SyntheticDF} to produce centroids. Details about the dictionary used in scene text recognition are shown in Scene text recognition experiment section. Soft distributions were derived from these embeddings and centroids, which we later refined. A threshold $T$ in Eq.(\ref{eq:thresh}) of 0.85 was applied to modify soft distribution probabilities. In Mask TextSpotterv3, we only applied our method to its spatial attention module, an auto-regressive-based model, for consistent comparison. Adam optimizer~\cite{Kingma2014AdamAM} was used for optimization.

\section{Experiments}

This section describes our method effectiveness on both scene text spotting (TotalText~\cite{Chng2017TotalTextAC}, SCUT-CTW1500 \cite{Liu2019CurvedST}, and ICDAR2015 \cite{Karatzas2015ICDAR2C}) and scene text recognition (IIIT5k~\cite{Mishra2009SceneTR}, SVT~\cite{Wang2011EndtoendST}, ICDAR2013~\cite{Karatzas2013ICDAR2R}, SVTP~\cite{Phan2013RecognizingTW}, ICDAR2015~\cite{Karatzas2015ICDAR2C}, CUTE80~\cite{Risnumawan2014ARA}) datasets. Besides, we will also describe experiment settings and datasets used for both pretraining and finetuning.
Figure~\ref{fig:self-correct} shows how our proposed method has improved baselines.

\subsection{Scene Text Spotting Experiments}

In our approach, modifying the underlying model architecture is unnecessary, making it non-mandatory to re-train the model with synthetic datasets. This implies that the existing pretrained model available in the baseline repositories can be employed. This aspect is especially important because re-training using extensive synthetic datasets often necessitates substantial computational resources and extended training durations. Empirical evidence suggests that our methodology substantially enhances the baseline by directly fine-tuning on target datasets, eliminating the need for re-train with the synthetic dataset. Furthermore, to investigate the feasibility of integrating linguistic insights during the pretraining phase, we re-trained ABCNetV2 using soft linguistic labels on Curved Synthetic, TotalText~\cite{Chng2017TotalTextAC}, ICDAR 15~\cite{Karatzas2015ICDAR2C}, ICDAR 13~\cite{Karatzas2013ICDAR2R}, MLT-2017~\cite{Nayef2017ICDAR2017RR}, and TextOCR~\cite{Singh2021TextOCRTL} datasets. We have termed this variant External Dataset (ED) and reported its outcomes with the results procured from direct fine-tuning on the provided checkpoints.

In our scene text spotting experiments, first, we removed all words present in the test set from a generic 90k dictionary. Next, we sampled a subset of words from this newly formed dictionary and merged it with the collection of words found in the training set to generate centroids. This step was implemented to ensure that the model would not unintentionally be exposed to words exclusive to the test set that had not appeared in the training set. For each dataset, we train a model using all the words from its training set. Subsequently, we use a single checkpoint to evaluate the model across all dictionaries associated with that dataset.

\vspace{-2mm}
\subsubsection{Experiment on Total-Text}

The Total-Text dataset~\cite{Chng2017TotalTextAC}, introduced in 2017, is a prominent benchmark for arbitrarily shaped scene text. Consisting of 1,555 photos (1,255 for training and 300 for testing), it features images with low-contrast backgrounds and complex texts. The dataset predominantly includes irregular text, with each photo containing at least one curved word. Text instances are annotated with polygons, and its extended version further annotates each instance with ten fixed points and a recognition sequence. The dataset exclusively contains English text.


\begin{table}[!tb]
\centering
\setlength{\tabcolsep}{4pt}
\scalebox{0.8}{\begin{tabular}{lcc}
\toprule
Method & None & Full \\
\midrule 
Charnet~\cite{Xing2019ConvolutionalCN} & 66.6 & - \\
SwinTextSpotter~\cite{Huang2022SwinTextSpotterST} & 74.3 & 84.1 \\
PAN++~\cite{Wang2021PANTE} & 68.6 & 78.6\\
GLASS~\cite{Ronen2022GLASSGT} & 76.6 & 83.0 \\
TTS~\cite{kittenplon2022towards} & 75.6 & 84.4\\
CRAFTS~\cite{Baek2020CharacterRA} & $\textbf{78.7}^{\textbf{(*)}}$ & - \\
SPTS~\cite{peng2022spts} & 74.2 & 82.4 \\
TESTR~\cite{Zhang2022TextST} & 73.3 & 83.9 \\
ABINet++~\cite{Fang2022ABINetAB} & 77.6 &\textbf{84.5} \\
\midrule
Mask TextSpotterv3~\cite{Liao2020MaskTV} & 71.2 & 78.4 \\
Mask TextSpotterv3+L \textbf{(Ours)} & \textbf{73.4} & \textbf{81.2} \\
\midrule
ABCNetv2 (github checkpoint) & 71.8 & 83.4 \\
ABCNetv2+L \textbf{(Ours)} & 74.5 & 84.6 \\
ABCNetv2+L (ED) \textbf{(Ours)} & \textbf{76.8} & $\textbf{87.1}^{\textbf{(*)}}$ \\
\bottomrule 
\end{tabular}}

\caption{{\bf Scene text spotting results on Total-Text}. The values shown in the table are H-mean scores for end-to-end models. \textit{None} and \textit{Full} represent without and with a dictionary, respectively; the dictionary contains all testing words in the inference phase. ED denotes for re-train with external data. Our methods significantly improved upon the baselines, ABCNetv2 and Mask TextSpotterv3, surpass ABINet++ when using a full dictionary with ABCNetv2+L (directly fine-tuning from provided checkpoint) and this improvement is even more significant when re-train with external data (ED). \textbf{(*)} denotes the best score. We report scores wherever they are available on paper or GitHub.
\label{tab:totaltext}}
\vspace{-5mm}
\end{table}

Table \ref{tab:totaltext} shows the performance of our ABCNetv2+L and Mask TextSpotterv3+L methods in both terms of using and not using the full dictionary during the inference process. The full dictionary includes all words appeared in test set. The ABCNetv2~\cite{Liu2022ABCNetVA} results in Table \ref{tab:totaltext} are the results reported by the authors in the paper. ABCNetv2~\cite{Liu2022ABCNetVA} (GitHub checkpoint) are the results of the checkpoint published by the authors on their GitHub. We use this implementation as the baseline for our proposal. Checkpoint results generally produce better results than the performance reported in the paper. We compare our results with reported and checkpoint-generated results to make a fair comparison. It can be seen that our proposal improved both ABCNetv2 and Mask TextSpotterv3 significantly in both scenarios with and without full dictionary. Moreover, ABCNetv2+L with dictionary and finetuned on provided checkpoint outperformed current state-of-the-art ABINet++~\cite{Fang2022ABINetAB} by 0.1. This improvement has been increased to 2.6 when doing re-train with ED using our method.

\newcommand\anncwith{\textwidth} 
\newcommand\anncheight{0.5\textwidth} 

\begin{figure}[!tbp] 
    \centering
    \begin{subfigure}[b]{0.48\textwidth} 
        \centering
        \includegraphics[width=\anncwith,height=\anncheight,keepaspectratio]{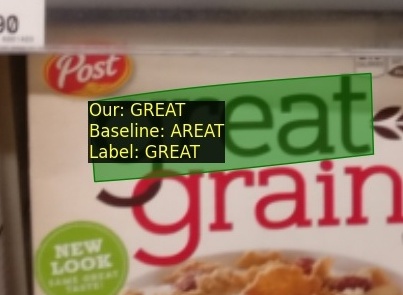}
        \caption{ABCNetv2: \hspace{2mm} \textcolor{red}{A}REAT\\
        \phantom{x}\hspace{2.5mm}ABCNetv2+L: GREAT}
    \end{subfigure}
    \hfill 
    \begin{subfigure}[b]{0.48\textwidth}
        \centering
        \includegraphics[width=\anncwith,height=\anncheight,keepaspectratio]{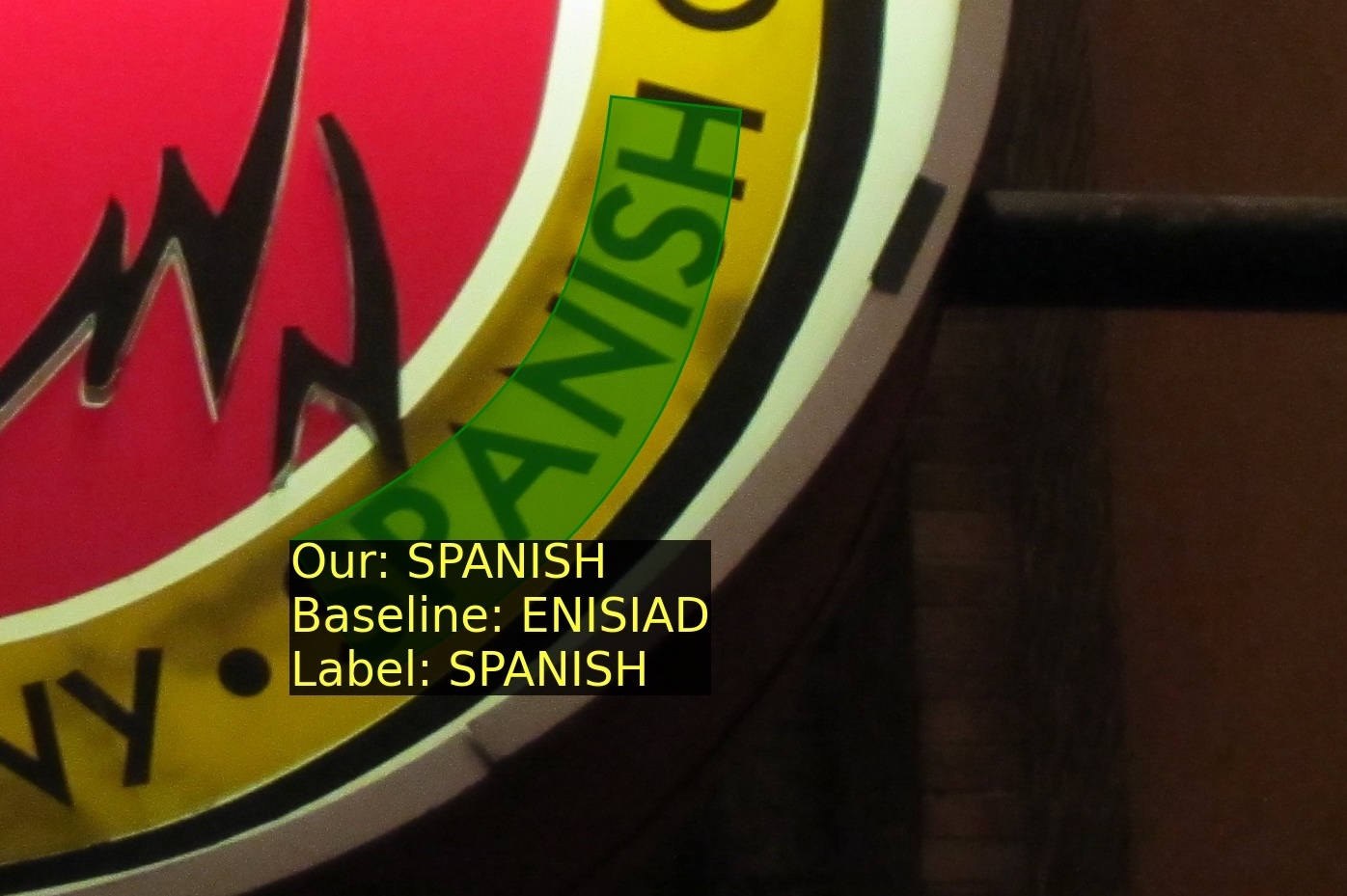}
        \caption{ABCNetv2: \hspace{2mm} \textcolor{red}{ENISIAD}\\
        \phantom{x}\hspace{2.5mm}ABCNetv2+L: SPANISH}
    \end{subfigure}
    \begin{subfigure}[b]{0.48\textwidth}
        \centering
        \includegraphics[width=\anncwith,height=\anncheight,keepaspectratio]{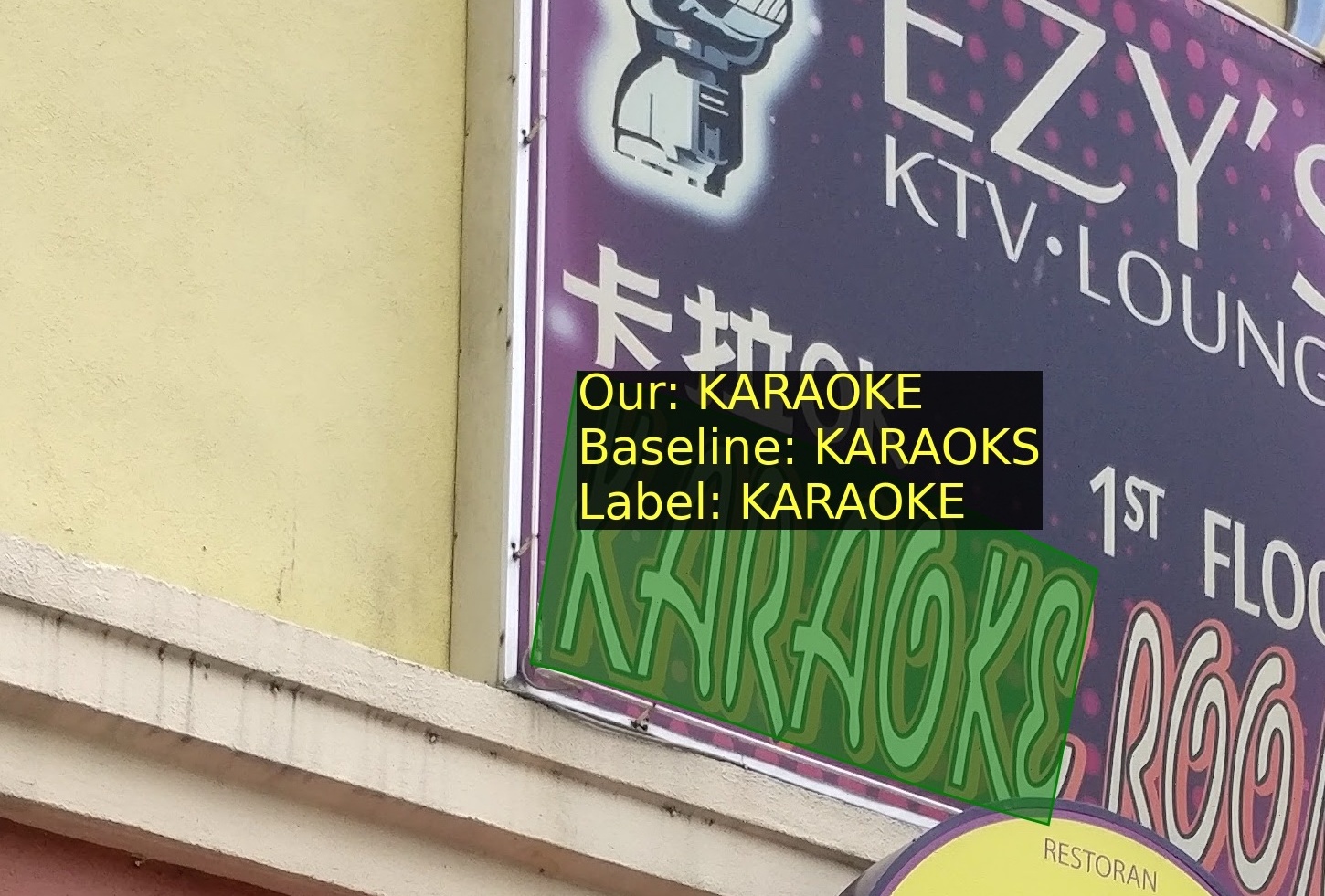}
        \caption{ABCNetv2: \hspace{2.3mm} KARAOK\textcolor{red}{S}\\
        \phantom{x}\hspace{2mm} ABCNetv2+L: KARAOKE}
    \end{subfigure}
    \hfill
    \begin{subfigure}[b]{0.48\textwidth}
        \centering
        \includegraphics[width=\anncwith,height=\anncheight,keepaspectratio]{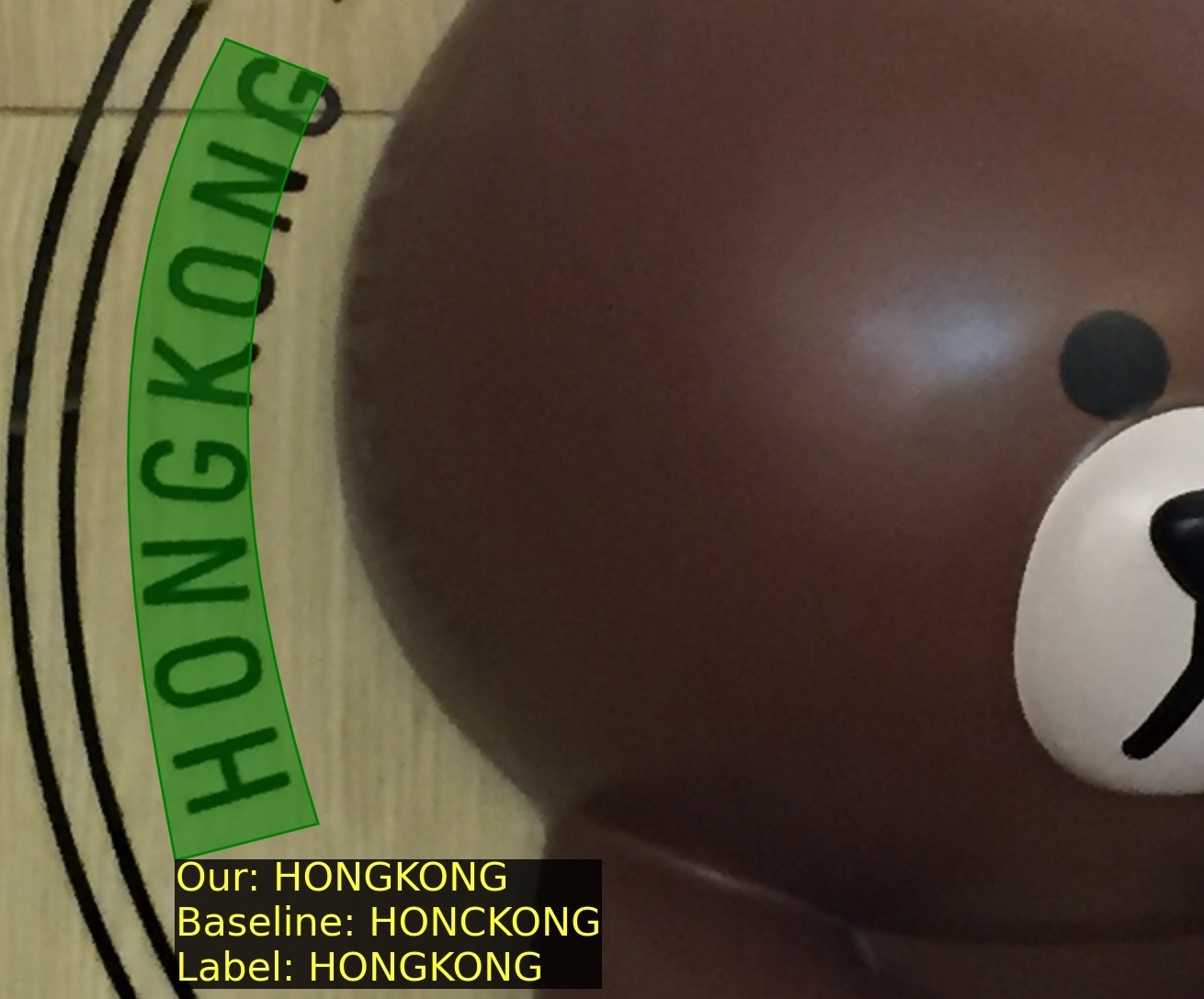}
        \caption{ABCNetv2: \hspace{1.3mm} HON\textcolor{red}{C}KONG\\
        \phantom{x}\hspace{1.1mm}ABCNetv2+L: HONGKONG}
    \end{subfigure}
    \caption{{\bf Qualitative results on Total-Text dataset.} Our approach is more capable of recognizing scene texts than the baseline. These outputs are directly taken from the model when the dictionary is not used in the testing phase.}
    \label{fig:self-correct}
\end{figure}

\subsubsection{Experiment on ICDAR 15}

Images from the ICDAR 15 ~\cite{Karatzas2015ICDAR2C} dataset were unintentionally taken from the real world through Google Glass. In contrast to earlier ICDAR 13 datasets~\cite{Karatzas2013ICDAR2R}, which featured cleanly captured text horizontally centered in the images, the ICDAR 15 dataset comprises 1000 training images and 500 testing photos that present significant challenges to text recognition algorithms. The images in the dataset feature complex backgrounds and text may appear in any orientation and at varying resolutions. Furthermore, the dataset is exclusively comprised of English text samples annotated at the word level.

Table \ref{tab:ic15} shows the experimental results of our proposal compared with previous works. For ICDAR 15, we use three scenarios to measure and evaluate the model, including strong, weak, and generic dictionary. Correspondingly, the strong dictionary for each image is a set of 100 words taken from the test set, in which all the words appearing in the image are in this group of 100 words. The weak dictionary is all the words that appear in the test set, and the generic vocabulary is the set of words that can appear, including a dictionary of 90k words \cite{Gupta2016SyntheticDF}. Our proposal effectively enhances the performance of the baseline models across all three types of dictionaries without any additional complexity. Notably, our ABCNetv2+L(ED) surpassed DeepSolo on strong and weak dictionaries by 0.3 and 0.5, and achieved state-of-the-art results on these dictionaries. Furthermore, while directly applying of our method to the pretrained model showed improvement, integrating it with external datasets (ED) in line with recent state-of-the-art elevated its performance remarkably.

\begin{table}[!tb]
    \centering
    \setlength{\tabcolsep}{2.5pt} 
    \caption{{\bf Scene text spotting results on ICDAR 15}. The values shown in the table are H-mean scores for end-to-end models. S, W, and G represent Strong, Weak, and Generic dictionaries used in the inference phase, respectively. Our method improved the baselines in all settings. Incorporating our method into ABCNetv2+L with ED, we outperformed current state-of-the-art on both Strong and Weak dictionary settings. \textbf{(*)} denotes the best score.}
    \label{tab:ic15}
    \small 
    \begin{tabular}{lccc}
    \toprule
    Method & S & W  & G \\
    \midrule 
    SwinTextSpotter~\cite{Huang2022SwinTextSpotterST} & 83.9 & 77.3 & 70.5\\
    PAN++~\cite{Wang2021PANTE} & 82.7 & 78.2 & 69.2\\
    GLASS~\cite{Ronen2022GLASSGT} & 84.7 & 80.1 & 76.3\\
    MANGO~\cite{Qiao2021MANGOAM} & 85.4 & 80.1 & 73.9\\
    CRAFTS~\cite{Baek2020CharacterRA} & 83.1 & 82.1 & 74.9 \\
    TTS~\cite{kittenplon2022towards} & 85.2 & 81.7 & 77.4 \\
    SPTS~\cite{peng2022spts} & 77.5 & 70.2 & 65.8 \\
    TESTR~\cite{Zhang2022TextST} & 85.2 & 79.4 & 73.6 \\
    ABINet++~\cite{Fang2022ABINetAB} & 86.1 & 81.9 & 77.8 \\
    DeepSolo~\cite{Ye2022DeepSoloLT} & \textbf{88.1} & \textbf{83.9} & $\textbf{79.5}^{\textbf{(*)}}$ \\
    \midrule 
    Mask Textspotterv3 ~\cite{Liao2020MaskTV} & 83.3 & 78.1 & 74.2 \\
    Mask TextSpotterv3+D~\cite{Nguyen2021DictionaryguidedST} & 85.2 & 81.9 &  75.9 \\
    Mask TextSpotterv3+L \textbf{(Ours)} & \textbf{85.9} & \textbf{81.9} & \textbf{77.4} \\
    \midrule 
    ABCNetv2 (github checkpoint) & 83.7 & 78.8 & 73.2 \\
    ABCNetv2+L \textbf{(Ours)} & 85.1 & 80.9 & 75.4 \\
    ABCNetv2+L(ED) \textbf{(Ours)} & $\textbf{88.4}^{\textbf{(*)}}$ & $\textbf{84.4}^{\textbf{(*)}}$ & \textbf{78.6} \\
    \bottomrule 
    \end{tabular}
\end{table}

\begin{table*}[ht]
\centering
\caption{\textbf{Comparison of scene text recognition accuracy on six datasets.} TargetDict denotes the list of words present in training sets of IIIT5k, SVT, IC13, IC15, SVTP, and CUTE datasets. The top-2 results are highlighted.
\label{tab:st-recognition}}

\begin{tabularx}{\linewidth}{
  >{\centering\arraybackslash}c 
  >{\centering\arraybackslash}m{2.5cm} 
  *{3}{>{\centering\arraybackslash}X}
  *{3}{>{\centering\arraybackslash}X}
}
\toprule
\multirow{2}{*}{\textbf{Method}} & \multirow{2}{*}{\textbf{Training Data}} & \multicolumn{3}{c}{\textbf{Regular Text}} & \multicolumn{3}{c}{\textbf{Irregular Text}} \\
\cmidrule(lr){3-5} \cmidrule(l){6-8}
& & IIIT & SVT & IC13 & SVTP & IC15 & CUTE \\
\midrule
ASTER~\cite{Shi2019ASTERAA} & MJ+ST & 93.4 & 89.5 & 91.8 & 78.5 & 76.1 & 79.5 \\
DAN~\cite{Wang2019DecoupledAN} & MJ+ST & 94.3 & 89.2 & 93.9 & 80.0 & 74.5 & 84.4 \\
RobustScanner~\cite{Yue2020RobustScannerDE} & MJ+ST & 95.3 & 88.1 & 94.8 & 79.5 & 77.1 & 90.3 \\
SAR~\cite{Li2018ShowAA} & ST+MJ & 91.5 & 84.5 & 91.0 & 76.4 & 69.2 & 83.3 \\
SEED~\cite{Qiao2020SEEDSE} & ST+MJ & 93.8 & 89.6 & 92.8 & 81.4 & 80.0 & 83.6 \\
ABINet~\cite{Fang2021ReadLH} & MJ+ST+Wiki & 96.2 & 93.5 & \textbf{97.4} & 89.3 & 86.0 & 89.2 \\
S-GTR~\cite{He2021VisualSA} & ST+MJ & 95.8 & 94.1 & 96.8 & 87.9 & 84.6 & 92.3 \\
LevOCR~\cite{Da2022LevenshteinO} & ST+MJ & 96.6 & 92.9 & 96.9 & 88.1 & 86.4 & 91.7 \\
SIGAT~\cite{Guan2022SelfsupervisedIG} & ST+MJ & 96.6 & \underline{95.1} & 96.8 & 90.5 & 83.0 & \underline{93.1} \\
MGP-STR~\cite{Wang2022MultiGranularityPF} & ST+MJ & 96.4 & 94.7 & \underline{97.3} & 91.0 & 87.2 & 90.3 \\
PARSeq~\cite{Bautista2022SceneTR} & ST+MJ & \textbf{97.0} & 93.6 & 96.2 & 82.9 & \textbf{88.9} & 92.2 \\
\midrule
CornerTransformer~\cite{Xie2022TowardUW} & ST+MJ & 95.9 & 94.6 & 96.4 & 91.5 & 86.3 & 92.0 \\
CornerTransformer+L\textbf{(Ours)} & ST+MJ & 96.4 & 95.0 & 96.9 & \underline{91.8} & 87.0 & 92.7 \\
CornerTransformer+L\textbf{(Ours)} & ST+MJ+TargetDict & \underline{96.7} & \textbf{95.4} & \textbf{97.4} & \textbf{92.2} & \underline{87.8} & \textbf{93.4} \\
\bottomrule
\end{tabularx}
\end{table*}

\subsubsection{Experiment on SCUT-CTW1500}


\vspace{-1mm}
The SCUT-CTW1500~\cite{Liu2019CurvedST} dataset, popular for arbitrary-shaped scene text, differs from Total-Text by including both Chinese and English. Annotations are at the text-line level, accommodating documents with stacked small text segments. It has 1000 training and 500 test images. For the SCUT-CTW1500 dataset, labels include word-level annotations and groups of text instances. Language knowledge is vital for accurate results, capturing both intra-word and inter-word relationships in long sequences. Table~\ref{tab:ctw1500} shows our ABCNetv2+L notably outperforms ABCNet2, especially with a dictionary during inference. Our method with ED surpasses all previous state-of-the-art models for SCUT-CTW1500 using a strong dictionary, highlighting the effectiveness and versatility of the proposed approach in addressing the challenges associated with localizing and recognizing scene text in the long line with arbitrary shapes.

\begin{table}[!tb]
\centering
\caption{{\bf Scene text spotting results on SCUT-CTW1500}. The values shown in the table are H-mean scores for end-to-end models. None and Strong represent without and with a strong dictionary in the inference phase, respectively. Our method improved the baselines in all settings and achieved state-of-the-art when evaluating without a dictionary for post-processing. \textbf{(*)} denotes the best score.}
\label{tab:ctw1500}
\setlength{\tabcolsep}{10pt}
\small 
\begin{tabular}{lcc}
\toprule
Method & None & Strong \\
\midrule 
TextDragon~\cite{Feng2019TextDragonAE} & 39.7 & 72.4 \\
MANGO~\cite{Qiao2021MANGOAM} & 58.9 & 78.7 \\
SwinTexSpotter~\cite{Huang2022SwinTextSpotterST} & 51.8 & 77.0 \\
Text Perceptron~\cite{Qiao2020TextPT} & 57.0 & - \\
ABINet++~\cite{Fang2022ABINetAB} & 60.2 & 80.3 \\
TESTR~\cite{Zhang2022TextST} & 56.0 & \textbf{81.5} \\
DeepSolo~\cite{Ye2022DeepSoloLT} & $\textbf{64.2}^{\textbf{(*)}}$ & 81.4 \\
\midrule
ABCNetv2~\cite{Liu2022ABCNetVA} & 57.5 & 77.2 \\
ABCNetv2+L \textbf{(Ours)} & 59.1 & 78.4 \\
ABCNetv2+L(ED) \textbf{(Ours)} & \textbf{61.2} & $\textbf{81.8}^{\textbf{(*)}}$ \\
\bottomrule 
\end{tabular}
\end{table}

\subsubsection{Detection Results}

Table~\ref{tab:detection} presents quantitative evidence of the effectiveness of our proposed method in enhancing not only text recognition but also text detection. In addition, Figure~\ref{fig:example} provides a visual representation of the qualitative improvements in text detection achieved through our approach. It is worth noting that unlike conventional object detection tasks, text detection is particularly sensitive since even the slightest inaccuracies in localizing words or including extraneous details can lead to erroneous recognition outcomes. To address this issue, we leverage language knowledge as a guiding signal to enhance the accuracy of our text detection model. Our experimental results, both quantitative and qualitative, clearly demonstrate the effectiveness of this approach in improving the overall quality of our text detection model. In summary, our simple yet effective method can significantly enhance text detection accuracy, which, in turn, directly influences text recognition outcomes.

\begin{table}[t]
    \small 
       \caption{Detection H-mean score comparison between ABCNetv2 and ABCNetv2+L. Our method improves detection performance on all three datasets.}
    \centering
    \begin{tabularx}{0.7\columnwidth}{Xccc}
        \toprule
        & TotalText & ICDAR15 & CTW1500 \\ [0.5ex] 
        \midrule 
        ABCNetv2 & 87.2 & 88.2 & 85.0 \\
        ABCNetv2+L (\textbf{Ours}) & 88.5 & 88.6 & 85.4 \\
        ABCNetv2+L (ED) (\textbf{Ours}) & \textbf{88.9} & \textbf{89.1} & \textbf{85.8} \\
        \bottomrule 
    \end{tabularx}
 
    \label{tab:detection}
\end{table}

\vspace{-2mm}
\subsection{Scene Text Recognition Experiments}
\label{recognition-experiment}
Our approach integrates linguistic knowledge into scene text models, enabling comprehensive generalization for text recognition. Using the CornerTransformer~\cite{Xie2022TowardUW} as a baseline, we train our model on SynthText and MJSynth datasets, following previous state-of-the-art methods. While image acquisition and labeling incur high costs, collecting domain-specific words is simpler. Our method capitalizes on this by generating centroids from in-domain text. We train the model with two centroid types: one using a 90k-word dictionary for fair comparison, and another incorporating words from training sets (excluding test-only words) of IIIT5k, SVT, IC13, IC15, SVTP, and CUTE. We call this list of in-domain words is  target dictionary (TargetDict).

The experimental results show in Table~\ref{tab:st-recognition} highlight the efficacy of our proposed method in enhancing the accuracy of the baseline model. This improvement is consistently observed across all six benchmark datasets when trained on MJSynth and SynthText datasets, following previous approaches. Compared with a method also incorporating linguistic knowledge into scene text recognition, ABINet, our method helps CornerTransformer surpass ABINet on IIIT5k and increases the existing improvement on SVT, SVTP, ICDAR 15, CUTE datasets more significantly. This superior performance is achieved without adding complexity to the model in training or inference.

Additionally, generating centroids with words from TargetDict increased the improvement by a large margin. Specially, CornerTransformer+L has improved the baseline's accuracy on IIT, SVT, ICDAR 13, SVTP, ICDAR 15, and CUTE by 0.8\%, 0.8\%, 1\%, 0.7\%, 1.5\%, and 1.4\%. Our method with TargetDict achieves state-of-the-art on SVT, ICDAR 13, SVTP, and CUTE, and secures the second-best results on ICDAR 15 and IIIT5k. These results underscore the model's effectiveness in bridging the domain gap between synthetic and real data. Importantly, our approach leverages target domain knowledge only from the text in target domain, minimizing the need for resource-intensive image collection and annotation processes.

\section{Conclusion}

We proposed a novel approach to transfer language knowledge to autoregressive-based scene text spotting models. Our approach overcomes ambiguous words using soft distributions based on character representation. We also proposed a process to generate soft distributions suitable for dictionaries without finetuning. Our approach was implemented on two scene text spotting backbones: ABCNetv2 and Mask TextSpotterv3, and one scene text recognition backbone: CornerTransformer, showing its generality. Extensive experiments on scene text spotting and recognition datasets can validate the effectiveness of our approach.

\bibliographystyle{unsrtnat}
\bibliography{main} 

\begin{thebibliography}{67}
\providecommand{\natexlab}[1]{#1}
\providecommand{\url}[1]{\texttt{#1}}
\expandafter\ifx\csname urlstyle\endcsname\relax
  \providecommand{\doi}[1]{doi: #1}\else
  \providecommand{\doi}{doi: \begingroup \urlstyle{rm}\Url}\fi

\bibitem[Jabnoun et~al.(2017)Jabnoun, Benzarti, and Amiri]{Jabnoun2017ANM}
Hanen Jabnoun, Faouzi Benzarti, and Hamid Amiri.
\newblock A new method for text detection and recognition in indoor scene for assisting blind people.
\newblock In \emph{International Conference on Machine Vision}, 2017.

\bibitem[Liu et~al.(2021)Liu, Xu, Li, Zhang, and Hou]{Liu2021ARO}
Shuhua Liu, Huixin Xu, Qi~Li, Fei Zhang, and Kun Hou.
\newblock A robot object recognition method based on scene text reading in home environments.
\newblock \emph{Sensors (Basel, Switzerland)}, 21, 2021.

\bibitem[Greenhalgh and Mirmehdi(2015)]{Greenhalgh2015RecognizingTT}
Jack Greenhalgh and Majid Mirmehdi.
\newblock Recognizing text-based traffic signs.
\newblock \emph{IEEE Transactions on Intelligent Transportation Systems}, 16:\penalty0 1360--1369, 2015.

\bibitem[Nguyen et~al.(2021)Nguyen, Nguyen, Tran, Tran, Ngo, Nguyen, and Hoai]{Nguyen2021DictionaryguidedST}
Nguyen~Le Nguyen, Thua Nguyen, Vinh~Quang Tran, Minh-Triet Tran, Thanh~Duc Ngo, Thien~Huu Nguyen, and Minh Hoai.
\newblock Dictionary-guided scene text recognition.
\newblock \emph{2021 IEEE/CVF Conference on Computer Vision and Pattern Recognition (CVPR)}, pages 7379--7388, 2021.

\bibitem[Baek et~al.(2019)Baek, Lee, Han, Yun, and Lee]{Baek2019CharacterRA}
Youngmin Baek, Bado Lee, Dongyoon Han, Sangdoo Yun, and Hwalsuk Lee.
\newblock Character region awareness for text detection.
\newblock \emph{2019 IEEE/CVF Conference on Computer Vision and Pattern Recognition (CVPR)}, pages 9357--9366, 2019.

\bibitem[Tang et~al.(2022)Tang, Qian, Song, Dong, Li, and Bai]{tang2022optimal}
Jingqun Tang, Wenming Qian, Luchuan Song, Xiena Dong, Lan Li, and Xiang Bai.
\newblock Optimal boxes: Boosting end-to-end scene text recognition by adjusting annotated bounding boxes via reinforcement learning.
\newblock In \emph{European Conference on Computer Vision}, pages 233--248. Springer, 2022.

\bibitem[Long et~al.(2018)Long, Ruan, Zhang, He, Wu, and Yao]{long2018textsnake}
Shangbang Long, Jiaqiang Ruan, Wenjie Zhang, Xin He, Wenhao Wu, and Cong Yao.
\newblock Textsnake: A flexible representation for detecting text of arbitrary shapes.
\newblock In \emph{Proceedings of the European conference on computer vision (ECCV)}, pages 20--36, 2018.

\bibitem[Zhang et~al.(2020)Zhang, Zhu, Hou, Liu, Yang, Wang, and Yin]{zhang2020deep}
Shi-Xue Zhang, Xiaobin Zhu, Jie-Bo Hou, Chang Liu, Chun Yang, Hongfa Wang, and Xu-Cheng Yin.
\newblock Deep relational reasoning graph network for arbitrary shape text detection.
\newblock In \emph{Proceedings of the IEEE/CVF Conference on Computer Vision and Pattern Recognition}, pages 9699--9708, 2020.

\bibitem[Zhu et~al.(2021)Zhu, Chen, Liang, Kuang, Jin, and Zhang]{zhu2021fourier}
Yiqin Zhu, Jianyong Chen, Lingyu Liang, Zhanghui Kuang, Lianwen Jin, and Wayne Zhang.
\newblock Fourier contour embedding for arbitrary-shaped text detection.
\newblock In \emph{Proceedings of the IEEE/CVF conference on computer vision and pattern recognition}, pages 3123--3131, 2021.

\bibitem[Liao et~al.(2020{\natexlab{a}})Liao, Wan, Yao, Chen, and Bai]{liao2020real}
Minghui Liao, Zhaoyi Wan, Cong Yao, Kai Chen, and Xiang Bai.
\newblock Real-time scene text detection with differentiable binarization.
\newblock In \emph{Proceedings of the AAAI conference on artificial intelligence}, volume~34, pages 11474--11481, 2020{\natexlab{a}}.

\bibitem[Liu et~al.(2019{\natexlab{a}})Liu, Liu, Sheng, Liang, Li, and Liu]{liu2019pyramid}
Jingchao Liu, Xuebo Liu, Jie Sheng, Ding Liang, Xin Li, and Qingjie Liu.
\newblock Pyramid mask text detector.
\newblock \emph{arXiv preprint arXiv:1903.11800}, 2019{\natexlab{a}}.

\bibitem[Wang et~al.(2019{\natexlab{a}})Wang, Liew, Zou, Zhou, and Feng]{wang2019panet}
Kaixin Wang, Jun~Hao Liew, Yingtian Zou, Daquan Zhou, and Jiashi Feng.
\newblock Panet: Few-shot image semantic segmentation with prototype alignment.
\newblock In \emph{Proceedings of the IEEE/CVF International Conference on Computer Vision}, pages 9197--9206, 2019{\natexlab{a}}.

\bibitem[Xie et~al.(2022)Xie, Fu, Zhang, Wang, and Bai]{Xie2022TowardUW}
Xudong Xie, Ling Fu, Zhifei Zhang, Zhaowen Wang, and Xiang Bai.
\newblock Toward understanding wordart: Corner-guided transformer for scene text recognition.
\newblock In \emph{European Conference on Computer Vision}, 2022.

\bibitem[Bautista and Atienza(2022)]{Bautista2022SceneTR}
Darwin Bautista and Rowel Atienza.
\newblock Scene text recognition with permuted autoregressive sequence models.
\newblock \emph{ArXiv}, abs/2207.06966, 2022.

\bibitem[Wang et~al.(2022)Wang, Da, and Yao]{Wang2022MultiGranularityPF}
P.~Wang, Cheng Da, and Cong Yao.
\newblock Multi-granularity prediction for scene text recognition.
\newblock In \emph{European Conference on Computer Vision}, 2022.

\bibitem[Zhao et~al.(2022)Zhao, Wu, Wu, Wilsbacher, and Wang]{Zhao2022BackgroundInsensitiveST}
Liang Zhao, Zhenyao Wu, Xinyi Wu, Greg Wilsbacher, and Song Wang.
\newblock Background-insensitive scene text recognition with text semantic segmentation.
\newblock In \emph{European Conference on Computer Vision}, 2022.

\bibitem[Liu et~al.(2022{\natexlab{a}})Liu, Yang, and Yin]{Liu2022OpenSetTR}
Chang Liu, Chun Yang, and Xu-Cheng Yin.
\newblock Open-set text recognition via character-context decoupling.
\newblock \emph{2022 IEEE/CVF Conference on Computer Vision and Pattern Recognition (CVPR)}, pages 4513--4522, 2022{\natexlab{a}}.

\bibitem[Bhunia et~al.(2021)Bhunia, Chowdhury, Sain, and Song]{Bhunia2021TowardsTU}
Ayan~Kumar Bhunia, Pinaki~Nath Chowdhury, Aneeshan Sain, and Yi-Zhe Song.
\newblock Towards the unseen: Iterative text recognition by distilling from errors.
\newblock \emph{2021 IEEE/CVF International Conference on Computer Vision (ICCV)}, pages 14930--14939, 2021.

\bibitem[Shi et~al.(2016{\natexlab{a}})Shi, Wang, Lyu, Yao, and Bai]{shi2016robust}
Baoguang Shi, Xinggang Wang, Pengyuan Lyu, Cong Yao, and Xiang Bai.
\newblock Robust scene text recognition with automatic rectification.
\newblock In \emph{Proceedings of the IEEE conference on computer vision and pattern recognition}, pages 4168--4176, 2016{\natexlab{a}}.

\bibitem[Wang and Hu(2017)]{wang2017gated}
Jianfeng Wang and Xiaolin Hu.
\newblock Gated recurrent convolution neural network for ocr.
\newblock \emph{Advances in Neural Information Processing Systems}, 30, 2017.

\bibitem[Cheng et~al.(2018)Cheng, Xu, Bai, Niu, Pu, and Zhou]{cheng2018aon}
Zhanzhan Cheng, Yangliu Xu, Fan Bai, Yi~Niu, Shiliang Pu, and Shuigeng Zhou.
\newblock Aon: Towards arbitrarily-oriented text recognition.
\newblock In \emph{Proceedings of the IEEE Conference on Computer Vision and Pattern Recognition}, pages 5571--5579, 2018.

\bibitem[Chng et~al.(2019)Chng, Liu, Sun, Ng, Luo, Ni, Fang, Zhang, Han, Ding, et~al.]{chng2019icdar2019}
Chee~Kheng Chng, Yuliang Liu, Yipeng Sun, Chun~Chet Ng, Canjie Luo, Zihan Ni, ChuanMing Fang, Shuaitao Zhang, Junyu Han, Errui Ding, et~al.
\newblock Icdar2019 robust reading challenge on arbitrary-shaped text-rrc-art.
\newblock In \emph{2019 International Conference on Document Analysis and Recognition (ICDAR)}, pages 1571--1576. IEEE, 2019.

\bibitem[Chng and Chan(2017)]{Chng2017TotalTextAC}
Chee-Kheng Chng and Chee~Seng Chan.
\newblock Total-text: A comprehensive dataset for scene text detection and recognition.
\newblock \emph{2017 14th IAPR International Conference on Document Analysis and Recognition (ICDAR)}, 01:\penalty0 935--942, 2017.

\bibitem[Liu et~al.(2019{\natexlab{b}})Liu, Jin, Zhang, Luo, and Zhang]{Liu2019CurvedST}
Yuliang Liu, Lianwen Jin, Shuaitao Zhang, Canjie Luo, and Sheng Zhang.
\newblock Curved scene text detection via transverse and longitudinal sequence connection.
\newblock \emph{Pattern Recognit.}, 90:\penalty0 337--345, 2019{\natexlab{b}}.

\bibitem[Karatzas et~al.(2015)Karatzas, i~Bigorda, Nicolaou, Ghosh, Bagdanov, Iwamura, Matas, Neumann, Chandrasekhar, Lu, Shafait, Uchida, and Valveny]{Karatzas2015ICDAR2C}
Dimosthenis Karatzas, Llu{\'i}s~G{\'o}mez i~Bigorda, Anguelos Nicolaou, Suman~K. Ghosh, Andrew~D. Bagdanov, M.~Iwamura, Jiri Matas, Luk{\'a}s Neumann, Vijay~Ramaseshan Chandrasekhar, Shijian Lu, Faisal Shafait, Seiichi Uchida, and Ernest Valveny.
\newblock Icdar 2015 competition on robust reading.
\newblock \emph{2015 13th International Conference on Document Analysis and Recognition (ICDAR)}, pages 1156--1160, 2015.

\bibitem[Su and Lu(2014)]{su2014accurate}
Bolan Su and Shijian Lu.
\newblock Accurate scene text recognition based on recurrent neural network.
\newblock In \emph{Asian conference on computer vision}, pages 35--48. Springer, 2014.

\bibitem[Shi et~al.(2016{\natexlab{b}})Shi, Bai, and Yao]{shi2016end}
Baoguang Shi, Xiang Bai, and Cong Yao.
\newblock An end-to-end trainable neural network for image-based sequence recognition and its application to scene text recognition.
\newblock \emph{IEEE transactions on pattern analysis and machine intelligence}, 39\penalty0 (11):\penalty0 2298--2304, 2016{\natexlab{b}}.

\bibitem[Liu et~al.(2016)Liu, Chen, Wong, Su, and Han]{liu2016star}
Wei Liu, Chaofeng Chen, Kwan-Yee~K Wong, Zhizhong Su, and Junyu Han.
\newblock Star-net: a spatial attention residue network for scene text recognition.
\newblock In \emph{BMVC}, volume~2, page~7, 2016.

\bibitem[Litman et~al.(2020)Litman, Anschel, Tsiper, Litman, Mazor, and Manmatha]{litman2020scatter}
Ron Litman, Oron Anschel, Shahar Tsiper, Roee Litman, Shai Mazor, and R~Manmatha.
\newblock Scatter: selective context attentional scene text recognizer.
\newblock In \emph{proceedings of the IEEE/CVF conference on computer vision and pattern recognition}, pages 11962--11972, 2020.

\bibitem[Liu et~al.(2018)Liu, Chen, and Wong]{liu2018char}
Wei Liu, Chaofeng Chen, and Kwan-Yee Wong.
\newblock Char-net: A character-aware neural network for distorted scene text recognition.
\newblock In \emph{Proceedings of the AAAI Conference on Artificial Intelligence}, volume~32, 2018.

\bibitem[Fang et~al.(2021)Fang, Xie, Wang, Mao, and Zhang]{Fang2021ReadLH}
Shancheng Fang, Hongtao Xie, Yuxin Wang, Zhendong Mao, and Yongdong Zhang.
\newblock Read like humans: Autonomous, bidirectional and iterative language modeling for scene text recognition.
\newblock \emph{2021 IEEE/CVF Conference on Computer Vision and Pattern Recognition (CVPR)}, pages 7094--7103, 2021.

\bibitem[Da et~al.(2022)Da, Wang, and Yao]{Da2022LevenshteinO}
Cheng Da, P.~Wang, and Cong Yao.
\newblock Levenshtein ocr.
\newblock \emph{ArXiv}, abs/2209.03594, 2022.

\bibitem[Wang et~al.(2021{\natexlab{a}})Wang, Xie, Fang, Wang, Zhu, and Zhang]{Wang2021FromTT}
Yuxin Wang, Hongtao Xie, Shancheng Fang, Jing Wang, Shenggao Zhu, and Yongdong Zhang.
\newblock From two to one: A new scene text recognizer with visual language modeling network.
\newblock \emph{2021 IEEE/CVF International Conference on Computer Vision (ICCV)}, pages 14174--14183, 2021{\natexlab{a}}.

\bibitem[Hinton et~al.(2015)Hinton, Vinyals, Dean, et~al.]{hinton2015distilling}
Geoffrey Hinton, Oriol Vinyals, Jeff Dean, et~al.
\newblock Distilling the knowledge in a neural network.
\newblock \emph{arXiv preprint arXiv:1503.02531}, 2\penalty0 (7), 2015.

\bibitem[Touvron et~al.(2021)Touvron, Cord, Douze, Massa, Sablayrolles, and J'egou]{Touvron2021TrainingDI}
Hugo Touvron, Matthieu Cord, Matthijs Douze, Francisco Massa, Alexandre Sablayrolles, and Herv'e J'egou.
\newblock Training data-efficient image transformers \& distillation through attention.
\newblock In \emph{ICML}, 2021.

\bibitem[Clark et~al.(2022)Clark, Garrette, Turc, and Wieting]{Clark2022CaninePA}
J.~Clark, Dan Garrette, Iulia Turc, and John Wieting.
\newblock Canine: Pre-training an efficient tokenization-free encoder for language representation.
\newblock \emph{Transactions of the Association for Computational Linguistics}, 10:\penalty0 73--91, 2022.

\bibitem[Gupta et~al.(2016)Gupta, Vedaldi, and Zisserman]{Gupta2016SyntheticDF}
Ankush Gupta, Andrea Vedaldi, and Andrew Zisserman.
\newblock Synthetic data for text localisation in natural images.
\newblock \emph{2016 IEEE Conference on Computer Vision and Pattern Recognition (CVPR)}, pages 2315--2324, 2016.

\bibitem[Kingma and Ba(2014)]{Kingma2014AdamAM}
Diederik~P. Kingma and Jimmy Ba.
\newblock Adam: A method for stochastic optimization.
\newblock \emph{CoRR}, abs/1412.6980, 2014.

\bibitem[Mishra et~al.(2009)Mishra, Karteek, and Jawahar]{Mishra2009SceneTR}
Anand Mishra, Alahari Karteek, and C.~V. Jawahar.
\newblock Scene text recognition using higher order language priors.
\newblock In \emph{British Machine Vision Conference}, 2009.
\newblock URL \url{https://api.semanticscholar.org/CorpusID:9695967}.

\bibitem[Wang et~al.(2011)Wang, Babenko, and Belongie]{Wang2011EndtoendST}
Kai Wang, Boris Babenko, and Serge~J. Belongie.
\newblock End-to-end scene text recognition.
\newblock \emph{2011 International Conference on Computer Vision}, pages 1457--1464, 2011.
\newblock URL \url{https://api.semanticscholar.org/CorpusID:14136313}.

\bibitem[Karatzas et~al.(2013)Karatzas, Shafait, Uchida, Iwamura, i~Bigorda, Mestre, Romeu, Mota, Almaz{\'a}n, and de~las Heras]{Karatzas2013ICDAR2R}
Dimosthenis Karatzas, Faisal Shafait, Seiichi Uchida, M.~Iwamura, Llu{\'i}s~G{\'o}mez i~Bigorda, Sergi~Robles Mestre, Joan~Mas Romeu, David~Fern{\'a}ndez Mota, Jon Almaz{\'a}n, and Llu{\'i}s-Pere de~las Heras.
\newblock Icdar 2013 robust reading competition.
\newblock \emph{2013 12th International Conference on Document Analysis and Recognition}, pages 1484--1493, 2013.

\bibitem[Phan et~al.(2013)Phan, Shivakumara, Tian, and Tan]{Phan2013RecognizingTW}
Trung~Quy Phan, Palaiahnakote Shivakumara, Shangxuan Tian, and Chew~Lim Tan.
\newblock Recognizing text with perspective distortion in natural scenes.
\newblock \emph{2013 IEEE International Conference on Computer Vision}, pages 569--576, 2013.
\newblock URL \url{https://api.semanticscholar.org/CorpusID:5619635}.

\bibitem[Risnumawan et~al.(2014)Risnumawan, Shivakumara, Chan, and Tan]{Risnumawan2014ARA}
Anhar Risnumawan, Palaiahnakote Shivakumara, Chee~Seng Chan, and Chew~Lim Tan.
\newblock A robust arbitrary text detection system for natural scene images.
\newblock \emph{Expert Syst. Appl.}, 41:\penalty0 8027--8048, 2014.
\newblock URL \url{https://api.semanticscholar.org/CorpusID:15559857}.

\bibitem[Nayef et~al.(2017)Nayef, Yin, Bizid, Choi, Feng, Karatzas, Luo, Pal, Rigaud, Chazalon, Khlif, Luqman, Burie, Liu, and Ogier]{Nayef2017ICDAR2017RR}
Nibal Nayef, Fei Yin, Imen Bizid, Hyunsoo Choi, Yuan Feng, Dimosthenis Karatzas, Zhenbo Luo, Umapada Pal, Christophe Rigaud, Joseph Chazalon, Wafa Khlif, Muhammad~Muzzamil Luqman, Jean-Christophe Burie, Cheng-Lin Liu, and Jean-Marc Ogier.
\newblock Icdar2017 robust reading challenge on multi-lingual scene text detection and script identification - rrc-mlt.
\newblock \emph{2017 14th IAPR International Conference on Document Analysis and Recognition (ICDAR)}, 01:\penalty0 1454--1459, 2017.
\newblock URL \url{https://api.semanticscholar.org/CorpusID:4761162}.

\bibitem[Singh et~al.(2021)Singh, Pang, Toh, Huang, Galuba, and Hassner]{Singh2021TextOCRTL}
Amanpreet Singh, Guan Pang, Mandy Toh, Jing Huang, Wojciech Galuba, and Tal Hassner.
\newblock Textocr: Towards large-scale end-to-end reasoning for arbitrary-shaped scene text.
\newblock \emph{2021 IEEE/CVF Conference on Computer Vision and Pattern Recognition (CVPR)}, pages 8798--8808, 2021.
\newblock URL \url{https://api.semanticscholar.org/CorpusID:234469662}.

\bibitem[Xing et~al.(2019)Xing, Tian, Huang, and Scott]{Xing2019ConvolutionalCN}
Linjie Xing, Zhi Tian, Weilin Huang, and Matthew~R. Scott.
\newblock Convolutional character networks.
\newblock \emph{2019 IEEE/CVF International Conference on Computer Vision (ICCV)}, pages 9125--9135, 2019.

\bibitem[Huang et~al.(2022)Huang, Liu, Peng, Liu, Lin, Zhu, Yuan, Ding, and Jin]{Huang2022SwinTextSpotterST}
Mingxin Huang, Yuliang Liu, Zhenghao Peng, Chongyu Liu, Dahua Lin, Shenggao Zhu, Nicholas~Jing Yuan, Kai Ding, and Lianwen Jin.
\newblock Swintextspotter: Scene text spotting via better synergy between text detection and text recognition.
\newblock \emph{2022 IEEE/CVF Conference on Computer Vision and Pattern Recognition (CVPR)}, pages 4583--4593, 2022.

\bibitem[Wang et~al.(2021{\natexlab{b}})Wang, Xie, Li, Liu, Liang, Zhibo, Lu, and Shen]{Wang2021PANTE}
Wenhai Wang, Enze Xie, Xiang Li, Xuebo Liu, Ding Liang, Yang Zhibo, Tong Lu, and Chunhua Shen.
\newblock Pan++: Towards efficient and accurate end-to-end spotting of arbitrarily-shaped text.
\newblock \emph{IEEE Transactions on Pattern Analysis and Machine Intelligence}, 44:\penalty0 5349--5367, 2021{\natexlab{b}}.

\bibitem[Ronen et~al.(2022)Ronen, Tsiper, Anschel, Lavi, Markovitz, and Manmatha]{Ronen2022GLASSGT}
Roi Ronen, Shahar Tsiper, Oron Anschel, Inbal Lavi, Amir Markovitz, and R.~Manmatha.
\newblock Glass: Global to local attention for scene-text spotting.
\newblock In \emph{European Conference on Computer Vision}, 2022.

\bibitem[Kittenplon et~al.(2022)Kittenplon, Lavi, Fogel, Bar, Manmatha, and Perona]{kittenplon2022towards}
Yair Kittenplon, Inbal Lavi, Sharon Fogel, Yarin Bar, R~Manmatha, and Pietro Perona.
\newblock Towards weakly-supervised text spotting using a multi-task transformer.
\newblock In \emph{Proceedings of the IEEE/CVF Conference on Computer Vision and Pattern Recognition}, pages 4604--4613, 2022.

\bibitem[Baek et~al.(2020)Baek, Shin, Baek, Park, Lee, Nam, and Lee]{Baek2020CharacterRA}
Youngmin Baek, Seung Shin, Jeonghun Baek, Sungrae Park, Junyeop Lee, Daehyun Nam, and Hwalsuk Lee.
\newblock Character region attention for text spotting.
\newblock \emph{ArXiv}, abs/2007.09629, 2020.

\bibitem[Peng et~al.(2022)Peng, Wang, Liu, Zhang, Huang, Lai, Li, Zhu, Lin, Shen, et~al.]{peng2022spts}
Dezhi Peng, Xinyu Wang, Yuliang Liu, Jiaxin Zhang, Mingxin Huang, Songxuan Lai, Jing Li, Shenggao Zhu, Dahua Lin, Chunhua Shen, et~al.
\newblock Spts: single-point text spotting.
\newblock In \emph{Proceedings of the 30th ACM International Conference on Multimedia}, pages 4272--4281, 2022.

\bibitem[Zhang et~al.(2022)Zhang, Su, Tripathi, and Tu]{Zhang2022TextST}
Xiang Zhang, Yongwen Su, Subarna Tripathi, and Zhuowen Tu.
\newblock Text spotting transformers.
\newblock \emph{2022 IEEE/CVF Conference on Computer Vision and Pattern Recognition (CVPR)}, pages 9509--9518, 2022.

\bibitem[Fang et~al.(2022)Fang, Mao, Xie, Wang, Yan, and Zhang]{Fang2022ABINetAB}
Shancheng Fang, Zhendong Mao, Hongtao Xie, Yuxin Wang, Chenggang~Clarence Yan, and Yongdong Zhang.
\newblock Abinet++: Autonomous, bidirectional and iterative language modeling for scene text spotting.
\newblock \emph{IEEE Transactions on Pattern Analysis and Machine Intelligence}, 45:\penalty0 7123--7141, 2022.

\bibitem[Liao et~al.(2020{\natexlab{b}})Liao, Pang, Huang, Hassner, and Bai]{Liao2020MaskTV}
Minghui Liao, Guan Pang, Jing Huang, Tal Hassner, and Xiang Bai.
\newblock Mask textspotter v3: Segmentation proposal network for robust scene text spotting.
\newblock \emph{ArXiv}, abs/2007.09482, 2020{\natexlab{b}}.

\bibitem[Liu et~al.(2022{\natexlab{b}})Liu, Shen, Jin, He, Chen, Liu, and Chen]{Liu2022ABCNetVA}
Yuliang Liu, Chunhua Shen, Lianwen Jin, Tong He, Peng Chen, Chongyu Liu, and Hao Chen.
\newblock Abcnet v2: Adaptive bezier-curve network for real-time end-to-end text spotting.
\newblock \emph{IEEE Transactions on Pattern Analysis and Machine Intelligence}, 44:\penalty0 8048--8064, 2022{\natexlab{b}}.

\bibitem[Qiao et~al.(2021)Qiao, Chen, Cheng, Xu, Niu, Pu, and Wu]{Qiao2021MANGOAM}
Liang Qiao, Ying Chen, Zhanzhan Cheng, Yunlu Xu, Yi~Niu, Shiliang Pu, and Fei Wu.
\newblock Mango: A mask attention guided one-stage scene text spotter.
\newblock In \emph{AAAI}, 2021.

\bibitem[Ye et~al.(2022)Ye, Zhang, Zhao, Liu, Liu, Du, and Tao]{Ye2022DeepSoloLT}
Maoyuan Ye, Jing Zhang, Shanshan Zhao, Juhua Liu, Tongliang Liu, Bo~Du, and Dacheng Tao.
\newblock Deepsolo: Let transformer decoder with explicit points solo for text spotting.
\newblock \emph{ArXiv}, abs/2305.19957, 2022.

\bibitem[Shi et~al.(2019)Shi, Yang, Wang, Lyu, Yao, and Bai]{Shi2019ASTERAA}
Baoguang Shi, Mingkun Yang, Xinggang Wang, Pengyuan Lyu, Cong Yao, and Xiang Bai.
\newblock Aster: An attentional scene text recognizer with flexible rectification.
\newblock \emph{IEEE Transactions on Pattern Analysis and Machine Intelligence}, 41:\penalty0 2035--2048, 2019.
\newblock URL \url{https://api.semanticscholar.org/CorpusID:206769320}.

\bibitem[Wang et~al.(2019{\natexlab{b}})Wang, Zhu, Jin, Luo, Chen, Wu, Wang, and Cai]{Wang2019DecoupledAN}
Tianwei Wang, Yuanzhi Zhu, Lianwen Jin, Canjie Luo, Xiaoxue Chen, Y.~Wu, Qianying Wang, and Mingxiang Cai.
\newblock Decoupled attention network for text recognition.
\newblock In \emph{AAAI Conference on Artificial Intelligence}, 2019{\natexlab{b}}.
\newblock URL \url{https://api.semanticscholar.org/CorpusID:209444482}.

\bibitem[Yue et~al.(2020)Yue, Kuang, Lin, Sun, and Zhang]{Yue2020RobustScannerDE}
Xiaoyu Yue, Zhanghui Kuang, Chenhao Lin, Hongbin Sun, and Wayne Zhang.
\newblock Robustscanner: Dynamically enhancing positional clues for robust text recognition.
\newblock In \emph{European Conference on Computer Vision}, 2020.
\newblock URL \url{https://api.semanticscholar.org/CorpusID:220525922}.

\bibitem[Li et~al.(2018)Li, Wang, Shen, and Zhang]{Li2018ShowAA}
Hui Li, Peng Wang, Chunhua Shen, and Guyu Zhang.
\newblock Show, attend and read: A simple and strong baseline for irregular text recognition.
\newblock \emph{ArXiv}, abs/1811.00751, 2018.
\newblock URL \url{https://api.semanticscholar.org/CorpusID:53301402}.

\bibitem[Qiao et~al.(2020{\natexlab{a}})Qiao, Zhou, Yang, Zhou, and Wang]{Qiao2020SEEDSE}
Zhi Qiao, Yu~Zhou, Dongbao Yang, Yucan Zhou, and Weiping Wang.
\newblock Seed: Semantics enhanced encoder-decoder framework for scene text recognition.
\newblock \emph{2020 IEEE/CVF Conference on Computer Vision and Pattern Recognition (CVPR)}, pages 13525--13534, 2020{\natexlab{a}}.
\newblock URL \url{https://api.semanticscholar.org/CorpusID:218862702}.

\bibitem[He et~al.(2021)He, Chen, Zhang, Liu, He, Wang, and Du]{He2021VisualSA}
Yang He, Chen Chen, Jing Zhang, Juhua Liu, Fengxiang He, Chaoyue Wang, and Bo~Du.
\newblock Visual semantics allow for textual reasoning better in scene text recognition.
\newblock In \emph{AAAI Conference on Artificial Intelligence}, 2021.
\newblock URL \url{https://api.semanticscholar.org/CorpusID:245502387}.

\bibitem[Guan et~al.(2022)Guan, Gu, Tu, Yang, Feng, Zhao, and Shen]{Guan2022SelfsupervisedIG}
Tongkun Guan, Chaochen Gu, Jingzheng Tu, Xuehang Yang, Qi~Feng, Yudi Zhao, and Wei Shen.
\newblock Self-supervised implicit glyph attention for text recognition.
\newblock 2022.
\newblock URL \url{https://api.semanticscholar.org/CorpusID:251589365}.

\bibitem[Feng et~al.(2019)Feng, He, Yin, Zhang, and Liu]{Feng2019TextDragonAE}
Wei Feng, Wenhao He, Fei Yin, Xu-Yao Zhang, and Cheng-Lin Liu.
\newblock Textdragon: An end-to-end framework for arbitrary shaped text spotting.
\newblock \emph{2019 IEEE/CVF International Conference on Computer Vision (ICCV)}, pages 9075--9084, 2019.

\bibitem[Qiao et~al.(2020{\natexlab{b}})Qiao, Tang, Cheng, Xu, Niu, Pu, and Wu]{Qiao2020TextPT}
Liang Qiao, Sanli Tang, Zhanzhan Cheng, Yunlu Xu, Yi~Niu, Shiliang Pu, and Fei Wu.
\newblock Text perceptron: Towards end-to-end arbitrary-shaped text spotting.
\newblock In \emph{AAAI Conference on Artificial Intelligence}, 2020{\natexlab{b}}.

\end{thebibliography}

\end{document}